\documentclass{article}

\usepackage{arxiv}

\usepackage[utf8]{inputenc} 
\usepackage[T1]{fontenc}    
\usepackage{hyperref}       
\usepackage{url}            
\usepackage{booktabs}       
\usepackage{amsfonts}       
\usepackage{nicefrac}       
\usepackage{microtype}      
\usepackage{lipsum}

\usepackage{times}
\usepackage{epsfig}
\usepackage{graphicx}
\usepackage{subfigure}
\usepackage{amsmath}
\usepackage{amssymb}
\usepackage{mathrsfs}
\usepackage{bm}
\usepackage{colortbl}
\usepackage{stfloats}
\usepackage{float}

\title{Extra Proximal-Gradient Inspired Non-local Network}

\author{
  Qingchao Zhang \\
  Mathematics\\
  University of Florida\\
  \texttt{qingchaozhang@ufl.edu} \\
   \And
 Yunmei Chen \\
  Mathematics\\
  University of Florida\\
  \texttt{yun@math.ufl.edu} \\
}

\begin{document}
\maketitle

\begin{abstract}
Variational method and deep learning method are two mainstream powerful approaches to solve inverse problems in computer vision. To take advantages of advanced optimization algorithms and powerful representation ability of deep neural networks, we propose a novel deep network for image reconstruction. The architecture of this network is inspired by our proposed accelerated extra proximal gradient algorithm. It is able to incorporate non-local operation to exploit the non-local self-similarity of the images and to learn the nonlinear transform, under which the solution is sparse. All the parameters in our network are learned from minimizing a loss function. Our experimental results show that our network outperforms several state-of-the-art deep networks with almost the same number of learnable parameter.

\end{abstract}

\section{Introduction}
Variational methods  have been among the most powerful tools for solving the following inverse problems in computer vision:
\begin{equation}\label{cp}
\min_\mathbf{x} f(\mathbf{x})+g (\mathbf{x}).
\end{equation}
  For image restoration and reconstruction,  $f(\mathbf{x})$ represents the data fidelity and $g(\mathbf{x})$ is a regularization functional exploiting the priors of $\mathbf{x}$. For instance, $f(\mathbf{x})=\frac{1}{2} \|A\mathbf{x}-\mathbf{y}\|^2_2$ and $g(\mathbf{x})=\|\mathbf{x}\|_1$ or $\|D\mathbf{x}\|_{2,1}$ (total variation of $\mathbf{x}$). Recent advance in optimization techniques has brought powerful tools to effectively solve this problem with strong theoretical convergence results. However,   the quality of the solutions highly depends on the sophisticated models for $f$ and $g$. Hence, the variational methods could suffer from the challenges of the flexibility.

Recently, learning based methods using deep neural network (DNN) have achieved impressive results in solving the inverse problem \eqref{cp} in computer vision. Due to the powerful representation ability \cite{Dong2014LearningAD} and universal approximation property of convolutional neural network (CNN)\cite{hornik1989multilayer}, the deep architecture of CNN can solve the inverse problem \eqref{cp} by learning from  training data, instead of hand-crafted functions.
However, training a deep neural network as a black box with limited insights for the solution requires large number of the training data to estimate millions or billions of parameters.
This is very expensive.  Moreover, they are usually sensitive to the specific problem.
 On the contrary, a typical variational model with an optimization solver can be applied to a class of problems.

To bridge the gap between variational methods and  deep learning  approaches and  take advantage of both, very recently,
an increasing number of works have contributed to the study of deep learning architectures inspired by variational models and optimization algorithms. The main idea of the work in this direction is combining a hand-crafted model $f$ with a learnable operator $g$ in the objective function in Eq. \eqref{cp}, then the solution is obtained by mapping an appropriate optimization algorithm to a deep neuron network. Each stage of the deep network corresponds to an iteration of the algorithm. The unknown operator $g$ at each iteration is represented by a set of parameters that are learned via minimizing a loss function. Roughly speaking,
 without a hand-crafted $g$, there have been two types of architectures of the  deep network inspired by optimization algorithms to solve Eq. \eqref{cp}. One of them is to learn the following proximal operator directly by viewing it as a CNN denoiser for $\mathbf{b}^k$ \cite{rick2017one, meinhardt2017learning, wang2016proximal, zhang2017learning}:
 \begin{equation} \label{prox}
Prox_{t_k,g}(\mathbf{b}^k)=:\mathop{\arg\min}_{\mathbf{x}} \frac{1}{2}\|\mathbf{x}-\mathbf{b}^k\|^2+t_k g(\mathbf{x}).
\end{equation}
The CNN denoiser is embedded into each iteration of an optimization algorithm to solve Eq. \eqref{cp}. For instance the CNN denoiser is placed into  the proximal gradient descent algorithm in \cite{meinhardt2017learning}, subproblem in half quadratic splitting in \cite{zhang2017learning}, subproblem in ADMM in \cite{rick2017one, meinhardt2017learning} and subproblem in primal-dual algorithm in \cite{wang2016proximal, meinhardt2017learning}. The other design is to learn the nonlinear function $g$ using CNN architecture, then to estimate the proximal operator \eqref{prox} with learned $g$.  For instance, the ADMM-Net \cite{NIPS2016_6406} maps the ADMM to a deep neural network with a $g$  of the form $g(\mathbf{x})=\sum_{i=1}^K \lambda_i h(D_i\mathbf{x})$, where the nonlinear mapping $h$ and linear operator $D$ are learned via the architecture of CNN using training data.
 The  ISTA-Net  \cite{Zhang2018ISTANetIO} is inspired by the iterative soft-thresholding algorithm (ISTA) to solve Eq. \eqref{cp} with a functional $g(\mathbf{x})$ of the form of $\|F(\mathbf{x})\|_1$, where the nonlinear mapping $F$ is designed as two convolution layers separated by a ReLU. The convolutional operators are linear operators, recently non-local operators have also been incorporated into $g(\mathbf{x})$ to build non-local networks for image restoration \cite{lefkimmiatis2017non, lefkimmiatis2018universal}, more detail will be recalled in Sec. \ref{2.2}.
The experimental results  in the literature have suggested strongly that the architectures of the deep neural networks inspired by optimization algorithms have great potential to improve the performance with much less parameters and enjoy better generalizability comparing with end-to-end learning deep networks.

  Motivated by the recent advances in  deep learning architectures relying on optimization algorithms, the aim of this work is to improve the performances of the existing DNNs by incorporating certain proven very powerful algorithms into the architecture of DNNs.  We will  present a novel deep network architecture  inspired by an accelerated extra-proximal gradient algorithm, which  is able to perform non-local operation and enhance sparsity of the the solution in a transformed domain.

\section{Related Work}\label{2}

In this section, we briefly review extragradient methods and non-local algorithms and networks.

\subsection{Extragradient Method}\label{2.1}
Extragradient method  first proposed by Korpelevich  \cite{Korpelevi1976An} has attracted much attention in optimization. It has been extended to solve variational inequality problems \cite{censor2011subgradient, Monteiro2011ComplexityOV}, constrained convex optimization \cite{luo1993error} and convex and non-convex composite optimization problems \cite{lin2017extragradient, nguyen2018extragradient} with theoretical performance guaranteed.
 Extragradient algorithms use an additional gradient step in a first order optimization algorithm to improve the convergence results. Intuitively, the additional gradient step implicitly takes into account the second order information of the objective function. The extragradient method might also be interpreted as an predictor-corrector mechanism in iterative optimization scheme to speed up the convergence. Hence we would like to incorporate the essence of this method to  the architecture of the deep network. The following two variants of the original extragradient algorithm are closely related to our design. The first one is the extended extragradient method in \cite{nguyen2018extragradient}. This algorithm uses extra proximal gradient steps at each iteration to solve non-convex composite minimization problem $\min_\mathbf{x} f(\mathbf{x})+g(\mathbf{x})$. Their scheme reads as
 \begin{equation} \label{extra1}
\mathbf{x}^{k+\frac{1}{2}}= Prox_{\alpha_k, g} (\mathbf{x}^k-\alpha_k\nabla f(\mathbf{x}^k)),
\end{equation}
\begin{equation} \label{extra2}
\mathbf{x}^{k+1} = Prox_{\beta_k, g} (\mathbf{x}^k-\beta_k\nabla f(\mathbf{x}^{k+\frac{1}{2}})).
\end{equation}
The second one is the accelerated  extragradient algorithm developed in  \cite{DiakonikolasO18}. This algorithm integrates Nesterov's accelerated gradient method for smooth convex optimization \cite{Nesterov:2014:ILC:2670022} into the extragradient method in \cite{Korpelevi1976An}. Different from the conventional extragradient method, this algorithm evaluates gradients in both steps at an interpolation of the previous two iterates rather than the previous iterate only.

\subsection{Non-local Method} \label{2.2}
Non-local methods for image restoration in variational methods \cite{buades2005non,buades2010image, dabov2007image, mairal2009non, elmoataz2008nonlocal, gilboa2008nonlocal, kindermann2005deblurring, lefkimmiatis2015nonlocal, zhou2005regularization}  and  non-variational approach such as notable BM3D algorithm \cite{dabov2007image} have significantly improved image quality, since the nonlocal operators based on patch model can exploit  the inherent non-local self-similarity property of images.
Recently, the success of the nonlocal methods has motivated the investigation of the architecture of DNNs that has the ability of
capturing long-range dependencies of the image \cite{lefkimmiatis2017non, lefkimmiatis2018universal, wang2018non}.
 The deep network architecture  for grayscale and color image denoising  in \cite{lefkimmiatis2017non} is inspired by the  projected gradient algorithm for solving a common variational image restoration model with a non-local regularization functional $g(\mathbf{x})=\sum_{r=1}^R\phi (L_r\mathbf{x})$. In $g(\mathbf{x})$ the gradient of the nonlinear function $\phi$ is represented by  Radial Basis Functions and non-local linear operators $L_r$ takes patch based model. But different from non-deep learning nonlocal models, the patch grouping is based on the similarity of the feature maps  rather than the intensities of the patches. The nonlocal operator  consists of three steps: (i). image patch extraction and the 2D patch transformation that can be combined in a convolutional layer, (ii). patch grouping that selects $K$ most similar patches, and (iii). collaborative filtering that performs a weighted sum of the $k$ transformed patches within the group in a  convolutional layer. In \cite{lefkimmiatis2018universal} the same nonlocal regularization functional as in \cite{lefkimmiatis2017non}  is employed into a deep network architecture to solve a common constrained  image restoration model. The experimental results showed the ability of this deep network on handling a wide range of noise levels using a single set of learned parameters.
  The nonlocal neural network proposed in \cite{wang2018non} can be viewed as a generalization of the classical non-local mean in \cite{buades2005non} that  computes the response  at a position as a weighted average of the image intensities  at all positions. Since in this model the weights depend only on the similarity of the image intensities between two positions, it is not robust to noise as the patch based non-local model.  The nonlocal neural network proposed in \cite{wang2018non} does not involve patches. It computes the response at a position as a weighted average of the feature maps  at all positions, and the weights
 depend on the similarity of the feature maps generated by convolutional operators.
   Hence, the weights implicitly depend on the feature maps in the patches with the size determined by the receptive fields.
\section{Architecture of  the Proposed Network}\label{sec:mainpart}

Motivated by  the recent advances in the deep learning architectures inspired optimization algorithms for image restoration and reconstruction, in this work we present  a novel deep neural network for solving Eq. \eqref{cp} with a  learnable transformed Lasso type regularization functional $\mathcal{G}( \mathbf{x} )$ to enhance the sparsity of the solution in $L_1$ norm of a transformed domain. That is
\begin{equation}\label{eq-000}
  \min_{\mathbf{x}} f(\mathbf{x}) + \lambda \| \mathcal{G}( \mathbf{x} )\|_1.
\end{equation}

The architecture of our proposed deep network is inspired by extra proximal gradient algorithm  with the
use of nonlocal operator  to exploit the inherent non-local self-similarity  for  the feature maps of the underlying images.
In Sec. \ref{Sec:extra}, we present an DNN architecture relying on an accelerated extra proximal gradient algorithm. In Sec. \ref{sec:Non-local}, we present how to combine local and non-local operation and integrate them into the extra proximal-gradient network presented in Sec. \ref{Sec:extra}.
Our proposed Extra Proximal-Gradient Non-local Network (EPN-Net) exhibit the advantage of the design relying on  an accelerated extra proximal gradient algorithm and of incorporating nonlocal operations.

 \subsection{Deep neural network inspired by an accelerated extra proximal gradient algorithm }\label{Sec:extra}

In this subsection we discuss the proposed architecture of the deep network inspired by an accelerated extra proximal gradient algorithm for solving Eq. \eqref{eq-000}. By integrating Nesterov's accelerated gradient method \cite{Nesterov} for minimizing smooth convex function

\begin{equation} \label{eq-15}
\widetilde{\mathbf{x}}^k =  \mathbf{x}^k + \gamma_k (\mathbf{x}^k - \mathbf{x}^{k-1}),
\end{equation}
\begin{equation}\label{eq-16}
  \mathbf{x}^{k+1} = \widetilde{\mathbf{x}}^k - \alpha \nabla f(\widetilde{\mathbf{x}}^k),
\end{equation}
into the  extra proximal gradient algorithm \cite{nguyen2018extragradient} in Eqs. \eqref{extra1}-\eqref{extra2},  we propose the following accelerated extra proximal gradient updating scheme:

\begin{figure*}[htb]
\begin{center}
   \includegraphics[width=1\linewidth]{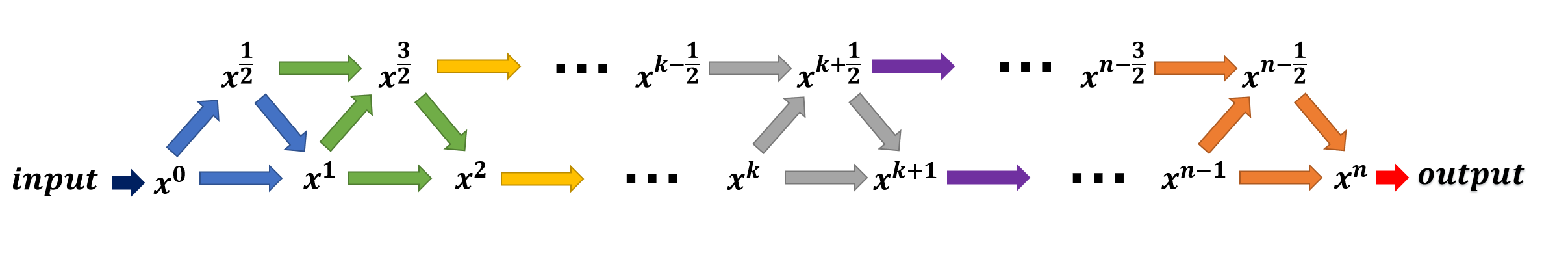}
\end{center}
   \caption{Illustration of our proposed EPN-Net architecture. The arrows in the same color imply they are in the same phase, i.e. they share the same operators and parameters $\mathcal{G}_k$, $\bm{\theta}_k$ and $\widetilde{\mathcal{G}}_k$ etc. In the $k^{th}$ phase, $\mathbf{x}^{k}$ and $\mathbf{x}^{k+1}$ are the input and output respectively. }
\label{fig-extragradient}
\end{figure*}



(i). In initial step select $\mathbf{x}^0$ (details in Sec. \ref{sec:experiment}) and set initial value $\gamma_0=0$.

(ii). For $k=1,2, \ldots$, we adopt the following iterations:
\begin{equation} \label{eq-17}
\widetilde{\mathbf{x}}^{k} = \mathbf{x}^{k} + \gamma_k (\mathbf{x}^{k} - \mathbf{x}^{k-\frac{1}{2}}),
\end{equation}
\begin{equation} \label{eq-17-2}
\mathbf{b}^{k+\frac{1}{2}} = \widetilde{\mathbf{x}}^{k}-\alpha_k \nabla f(\widetilde{\mathbf{x}}^{k}),
\end{equation}
\begin{equation} \label{eq-18}
\mathbf{x}^{k+\frac{1}{2}} = Prox_{\lambda\alpha_k, \|\mathcal{G}_k\|_1}(\mathbf{b}^{k+\frac{1}{2}}),
\end{equation}
\begin{equation} \label{eq-19}
\widehat{\mathbf{x}}^{k} = \mathbf{x}^{k+\frac{1}{2}} + \gamma_k (\mathbf{x}^{k+\frac{1}{2}} - \mathbf{x}^{k}),
\end{equation}
\begin{equation} \label{eq-19-2}
\mathbf{b}^{k+1} = \widehat{\mathbf{x}}^{k}-\beta_k \nabla f(\widehat{\mathbf{x}}^{k}),
\end{equation}
\begin{equation}\label{eq-20}
  \mathbf{x}^{k+1} = Prox_{\lambda\beta_k, \|\mathcal{G}_k\|_1} (\mathbf{b}^{k+1}),
\end{equation}
where $\alpha_k$ and $\beta_k$ are step sizes in the $k^{th}$ updating scheme. In our implementation, we adopt equal step sizes ($\alpha_k=\beta_k$). Here, each updating scheme has its own $\mathcal{G}_k$ to increase capacity.

Motivated by residual shortcut in ResNet \cite{ResNet} which has been proven to boost convergence and enhance accuracy, at each iteration $k$, we model
\begin{equation}\label{res-1}
  \mathbf{x}^{k+\frac{1}{2}} = \mathbf{b}^{k+\frac{1}{2}} + \mathbf{r}_{k+\frac{1}{2}} (\mathbf{x}^{k+\frac{1}{2}}),
\end{equation}
similarly,
\begin{equation}\label{res-2}
  \mathbf{x}^{k+1} = \mathbf{b}^{k+1} + \mathbf{r}_{k+1} (\mathbf{x}^{k+1}),
\end{equation}
where $\mathbf{r}_k(\mathbf{x}^k)$ denotes residual of $\mathbf{x}^k$.

Instead of updating the $\mathbf{x}^{k+\frac{1}{2}}$ and $\mathbf{x}^{k+1}$ directly as in Eqs. \eqref{eq-18} and \eqref{eq-20}, we learn the residual $\mathbf{r}_{k+\frac{1}{2}}(\mathbf{x}^{k+\frac{1}{2}})$ and $\mathbf{r}_{k+1}(\mathbf{x}^{k+1})$ first. Then we can easily get $\mathbf{x}^{k+\frac{1}{2}}$ and $\mathbf{x}^{k+1}$ from Eqs. \eqref{res-1} and \eqref{res-2}.
To this end we propose that the residual $\mathbf{r}_{k+\frac{1}{2}}(\mathbf{x}^{k+\frac{1}{2}})$ and $\mathbf{r}_{k+1}(\mathbf{x}^{k+1})$ take the form
\begin{equation}\label{r-1}
  \mathbf{r}_{k+\frac{1}{2}}(\mathbf{x}^{k+\frac{1}{2}}) = \widetilde{\mathcal{G}}_k \circ \mathcal{G}_k(\mathbf{x}^{k+\frac{1}{2}}),
\end{equation}
\begin{equation}\label{r-2}
  \mathbf{r}_{k+1}(\mathbf{x}^{k+1}) = \widetilde{\mathcal{G}}_k \circ \mathcal{G}_k(\mathbf{x}^{k+1}),
\end{equation}
where $\widetilde{\mathcal{G}}_k$ and ${\mathcal{G}}_k$ are learnable nonlinear transformation, $\mathbf{x}^{k+\frac{1}{2}}$ and $\mathbf{x}^{k+1}$ are sparse under transformation $\mathcal{G}_k$. Below we present our design of learnable nonlinear operator $\mathcal{G}_k$ and $\widetilde{\mathcal{G}}_k$:

\textbf{Nonlinear operator $\mathcal{G}_k$:}
We design the nonlinear operator $\mathcal{G}_k$ consisting of two linear convolutional operators $\mathbf{A}$, $\mathbf{B}$ separated by a ReLU and one linear feature generator $\mathbf{D}$.
Specifically, $\mathcal{G}_k(\mathbf{x})=\mathbf{B} ReLU (\mathbf{A}\mathbf{D}\mathbf{x})$.
$\mathcal{G}_k$ has local property because of the local property of convolution, in CNN the output of a neuron is influenced by a local region in original image \cite{LeB17}.
This effective local region is called receptive field (RF), whose size is mainly decided by convolution kernel size, network depth and pooling \cite{NIPS2016_6203}.
Here, we adopt $\mathbf{A}$, $\mathbf{B}$ and $\mathbf{D}$ as $3 \times 3$ conv layer.
 Thus, the RF of $\mathcal{G}_k$ is $7 \times 7$ \cite{vgg}, which means every neuron at the output of $\mathcal{G}_k$ covers a neighboring $7 \times 7$ patch in original image.


  According to \textbf{Theorem $\mathbf{1}$} in ISTA-Net$^+$ \cite{Zhang2018ISTANetIO}, if operator $\mathcal{G}_k$ has the format $\mathcal{B} ReLU (\mathcal{A})$ (where $\mathcal{A}$ and $\mathcal{B}$ are linear operators), $\mathcal{G}_k$ will satisfy $\mathbb{E}[\|\mathcal{G}_k(\mathbf{x})-\mathcal{G}_k({\mathbf{b}})\|^2_2] = \tau \mathbb{E}[\|\mathbf{x}-{\mathbf{b}}\|^2_2]$, $\tau$ is a constant depending on $\mathcal{A}$ and $\mathcal{B}$.
  Our designed $\mathcal{G}_k$ satisfies the format since $\mathcal{A} = \mathbf{A}\mathbf{D}$ and $\mathcal{B} = \mathbf{B}$ are both linear operators.
  To have $\mathcal{G}_k (\mathbf{x}^{k+\frac{1}{2}})$ and $\mathcal{G}_k (\mathbf{x}^{k+1})$ sparse, we let $\mathbf{z}^k= \mathcal{G}_k( \overline{ \mathbf{x}}^k)$ to be the minimizer of the Lasso problem:
\begin{equation}\label{sparse-2}
 \mathbf{z}^k= \mathop{\arg\min}_{\mathbf{z}} \frac{1}{2}\|\mathbf{z}-\mathcal{G}_k(\overline{\mathbf{b}}^k)\|^2_2+\theta_k \|\mathbf{z}\|_1,
\end{equation}
where $\overline{ \mathbf{x}}^k$ and $\overline{ \mathbf{b}}^k$ denote $\mathbf{x}^{k+\frac{1}{2}}$, $\mathbf{x}^{k+1}$ and $\mathbf{b}^{k+\frac{1}{2}}$, $\mathbf{b}^{k+1}$ for simplicity;
$\theta_k = \tau \lambda \alpha_k$ or $\tau \lambda \beta_k$.

Convolutional operator $\mathbf{B}$ of $\mathcal{G}_k$ corresponds to $N_f$ kernels, hence $\mathbf{z}^k$ has $N_f$ channels ($N_f$ is set to $32$ by default in our network). The iterative shrinkage thresholding algorithm (ISTA) is applied channel-wise for these multi-channel feature maps to get
\begin{equation}\label{sparse-3}
 \mathcal{G}_k(\overline{ \mathbf{x}}^k) =  \mathbf{z}^k = soft(\mathcal{G}_k(\overline{\mathbf{b}}^k), \bm{\theta}_k).
\end{equation}
Each channel has its own threshold for shrinkage operator. We use $\bm{\theta}_k =\{{\theta}_k^1, {\theta}_k^2, ..., {\theta}_k^{N_f}\}$ to present the thresholds for all $N_f$ channels.
Here $soft(\cdot, \bm{\theta}_k)$ denotes shrinkage operator with learnable thresholds $\bm{\theta}_k$.



\textbf{Nonlinear operator $\widetilde{\mathcal{G}}_k$:}
The operator $\widetilde{\mathcal{G}}_k$ is to generate and combine the feature maps for $\mathcal{G}_k(\mathbf{x}^{k+\frac{1}{2}})$ and $\mathcal{G}_k(\mathbf{x}^{k+1})$ to finally output residual $\mathbf{r}_{k+\frac{1}{2}}$ and $\mathbf{r}_{k+1}$.
Motivated by the symmetrical structure of U-Net \cite{DBLP:journals/corr/RonnebergerFB15}, we model $\widetilde{\mathcal{G}}_k(\mathbf{x})$ as $\widetilde{\mathbf{D}} \widetilde{\mathbf{A}} ReLU (\widetilde{\mathbf{B}}\mathbf{x})$, where $\widetilde{\mathbf{A}}$, $\widetilde{\mathbf{B}}$ and $\widetilde{\mathbf{D}}$ are all $3\times3$ convolutional operators.

Based upon above design of ${\mathcal{G}}_k$ and $\widetilde{\mathcal{G}}_k$, and employing residual method in Eqs. \eqref{res-1}\eqref{res-2}, we improve the updating scheme in Eqs. \eqref{eq-18}\eqref{eq-20} to the following:
\begin{equation} \label{eq-23}
\mathbf{x}^{k+\frac{1}{2}} = \mathbf{b}^{k+\frac{1}{2}} + \widetilde{\mathcal{G}}_k \circ soft(\mathcal{G}_k(\mathbf{b}^{k+\frac{1}{2}}), \bm{\theta}_k),
\end{equation}
\begin{equation}\label{eq-26}
  \mathbf{x}^{k+1} =  \mathbf{b}^{k+1} + \widetilde{\mathcal{G}}_k \circ soft(\mathcal{G}_k( \mathbf{b}^{k+1}), \bm{\theta}_k).
\end{equation}

The above updating scheme boosts the sparsity of $\mathbf{x}^{k+\frac{1}{2}}$ and $\mathbf{x}^{k+1}$ under the learnable nonlinear transformation $\mathcal{G}_k$.

%

The network inspired by above algorithm computes intermediate variable $\mathbf{x}^{k+\frac{1}{2}}$ at each updating scheme, as depicted in Fig. \ref{fig-extragradient}.
It is different from simply adding more layers. In this network the updating steps for $\mathbf{x}^{k+\frac{1}{2}}$ and $\mathbf{x}^{k+1}$ use the same network operators $\mathcal{G}_k$, $\widetilde{\mathcal{G}}_k$ and parameters $\bm{\theta}_k$, it can be viewed as predictor-corrector mechanism for updating $\mathbf{x}^{k+1}$.
As the architecture shown in Fig. \ref{fig-extragradient}, the arrows with the same color represent the updating schemes sharing the same set of learnable parameters.
Because parameters are shared in each stage, it will not increase the number of learnable parameters and will not extend the depth of the network. 
\begin{figure*}[htb]
\begin{center}
   \includegraphics[width=0.95\linewidth]{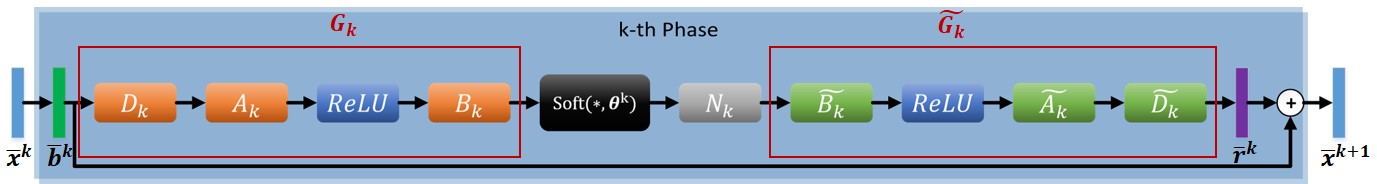}
\end{center}
   \caption{Illustration of the $k^{th}$ phase of our proposed EPN-Net.
Here, "$\bigoplus$" represents element-wise sum. All operators $\mathbf{A}_k$, $\mathbf{B}_k$, $\mathbf{D}_k$ and $\widetilde{\mathbf{A}}_k$, $\widetilde{\mathbf{B}}_k$, $\widetilde{\mathbf{D}}_k$ denote $3\times3$ convolutions; $\overline{\mathbf{x}}^{k+1}$ represents the output $\mathbf{x}^{k+\frac{1}{2}}$ and $\mathbf{x}^{k+1}$ in Eqs. \eqref{eq-23-2} and \eqref{eq-26-2}.
}
\label{fig-1}
\end{figure*}
\subsection{Non-local Operation} \label{sec:Non-local}

In Sec. \ref{Sec:extra}, both of the learnable transformations $\mathcal{G}_k$ and $\widetilde{\mathcal{G}}_k$ are local operators composed of convolutions followed by ReLUs.
In order to exploit the non-local self-similarity property of the feature maps of the underlying image, we propose to integrate local and non-local \cite{wang2018non} operators in each updating scheme.
By inserting non-local operator $\mathcal{N}_k$ to the updating scheme for $\overline {\mathbf{r}}_{k}(\overline{\mathbf{x}}^{k}) = \widetilde{\mathcal{G}}_k \circ \mathcal{G}_k(\overline{\mathbf{x}}^{k})$ in Eqs. \eqref{r-1} and \eqref{r-2}, in this section we propose to learn the residue $\overline {\mathbf{r}}_{k}(\overline{\mathbf{x}}^{k})$ with the structure $\overline {\mathbf{r}}_{k}(\overline{\mathbf{x}}^{k}) = \widetilde{\mathcal{G}}_k \circ \mathcal{N}_k\circ \mathcal{G}_k(\overline{\mathbf{x}}^{k})$, where $\overline {\mathbf{r}}_{k}$ denotes $\mathbf{r}_{k+\frac{1}{2}}$ and $\mathbf{r}_{k+1}$ for simplicity.
Specifically, from Eq. \eqref{sparse-3} we get:
\begin{equation}\label{eq-26-0000}
  \overline{{\mathbf{r}}}_{k}(\overline{\mathbf{x}}^{k})  = \widetilde{\mathcal{G}}_k\circ \mathcal{N}_k(soft(\mathcal{G}_k(\overline{\mathbf{b}}^{k}), \bm{\theta}_k)).
\end{equation}
Upon this we model updating schemes in Eqs. \eqref{eq-23} \eqref{eq-26} as following:
\begin{equation} \label{eq-23-2}
\mathbf{x}^{k+\frac{1}{2}} = \mathbf{b}^{k+\frac{1}{2}} + \widetilde{\mathcal{G}}_k\circ \mathcal{N}_k(soft(\mathcal{G}_k(\mathbf{b}^{k+\frac{1}{2}}), \bm{\theta}_k)),
\end{equation}
\begin{equation}\label{eq-26-2}
  \mathbf{x}^{k+1} =  \mathbf{b}^{k+1} + \widetilde{\mathcal{G}}_k\circ \mathcal{N}_k(soft(\mathcal{G}_k(\mathbf{b}^{k+1}), \bm{\theta}_k)).
\end{equation}

The detailed design for the non-local operator $\mathcal{N}_k$ is given in the following text.

 \textbf{Non-local operator $\mathcal{N}_k$:}
 To maximize the use of the information from the feature maps of the underlying image, our nonlocal operator $\mathcal{N}_k$ is a nonlinear map that  combines the local nonlinear map $\mathcal{G}_k(\overline{ \mathbf{x}}^k)$ in Eq. \eqref{sparse-3} and a non-local block $\mathcal{M}_k(\overline{ \mathbf{x}}^k)$ which computes the nonlocal mean of $\mathcal{G}_k(\overline{ \mathbf{x}}^k)$.

\begin{figure}[htp]
\begin{center}
   \includegraphics[width=0.5\linewidth]{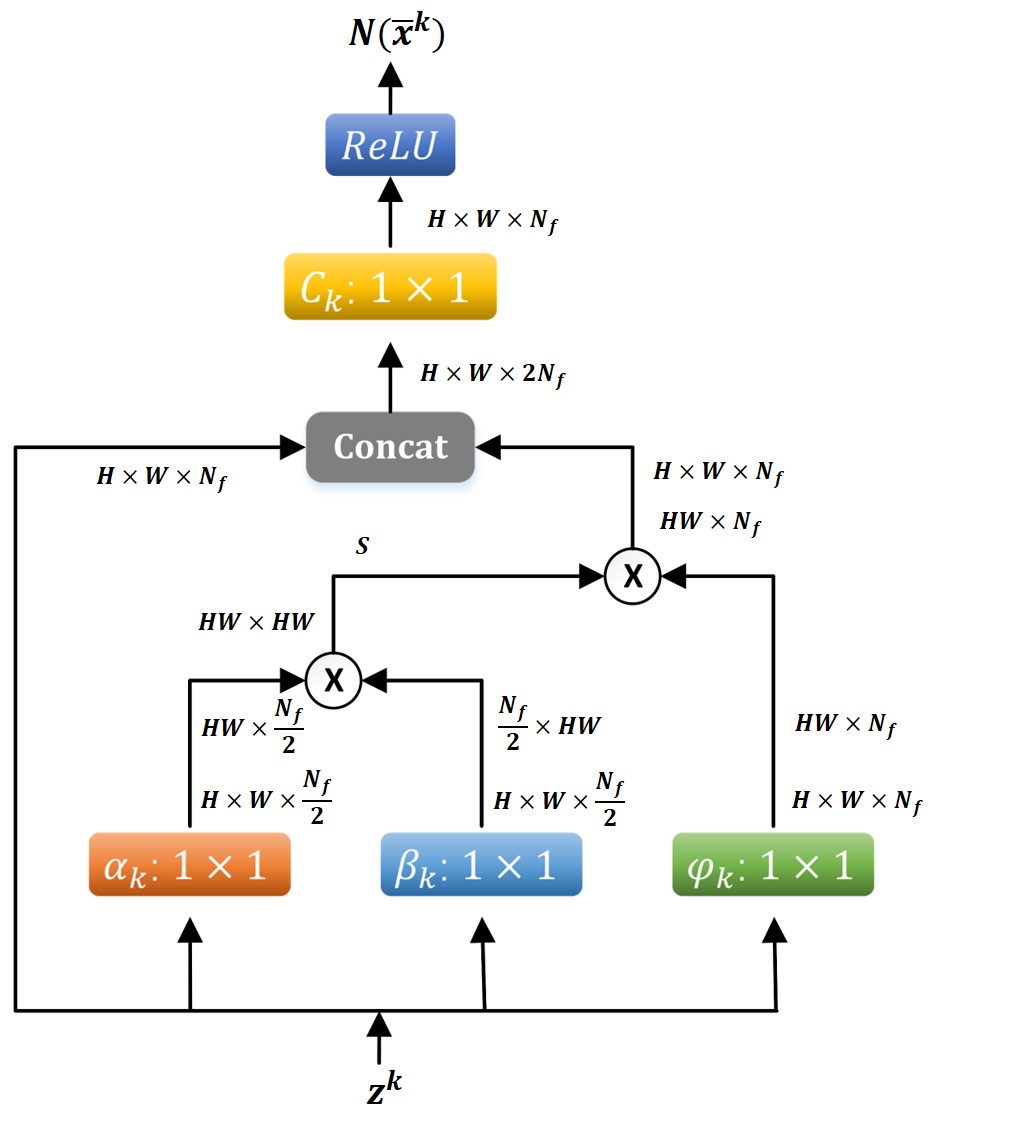}
\end{center}
   \caption{Illustration of Non-local operator. "$\bigotimes$" denotes matrix multiplication, $S$ denotes $softmax$ opertion. Both input $\mathbf{z}^k$ and output $\mathcal{N}(\overline{\mathbf{x}}^k)$ are of the same shape $H \times W\times N_f$. }
\label{fig-nonlocal}
\end{figure}

  \begin{itemize}
  \item\emph{Non-local Block $\mathcal{M}_k(\overline{ \mathbf{x}}^k)$}:
  The design of $\mathcal{M}_k(\overline{ \mathbf{x}}^k)$ is adopted from the nonlocal block in \cite{wang2018non} which computes a weighted average of features at all positions. More precisely, denote by $\mathbf{z}_j$ the input vector at position $j$ and $\mathbf{v}_i$ the response vector at position $i$, the nonlocal block in \cite{wang2018non} computes $\mathbf{v}_i$ by:
\begin{equation}\label{eq-11}
   \mathbf{v}_i = \sum_{\forall j}\bm{\omega}_{i,j} \cdot \varphi(\mathbf{z}_j),
 \end{equation}
 where the unary function $\varphi$ computes a representation of the input signal at the position $j$, and $\bm{\omega}_{i,j}$ is the normalized weights depending on the similarity between $\mathbf{z}_i$ and $\mathbf{z}_j$. The mapping $\varphi$ corresponds to a learnable matrix $\mathbf{W}_\varphi$ (implemented as $1 \times 1$ convolution).

 %
Several possible choices of the weights are given in \cite{wang2018non}. In this work we set it to be Embedded Gaussian \cite{wang2018non} with bottleneck:
  \begin{equation}\label{eq-11-0}
   \bm{\omega}_{i,j} = \frac{e^{\alpha(\mathbf{z}_i)^T \beta(\mathbf{z}_j)}}{\sum_{\forall j}e^{\alpha(\mathbf{z}_i)^T \beta(\mathbf{z}_j)}},
 \end{equation}
 where $\alpha = \mathbf{W}_\alpha$ and $\beta = \mathbf{W}_\beta$ respectively. In implementation, they correspond to ${N_f}/{2}$ convolutional filters of kernel size $3\times3$. We employ this bottleneck structure to reduce computation.



In addition, Eqs. \eqref{eq-11}\eqref{eq-11-0} can be implemented efficiently by matrix multiplication:
 \begin{equation}\label{eq-11-1}
   \mathbf{v}^k = \mathcal{S}((\mathbf{W}_\alpha^k \mathbf{z}^k)^\mathrm{T} \mathbf{W}_\beta^k \mathbf{z}^k)\mathbf{W}_\varphi^k \mathbf{z}^k,
 \end{equation}
 where $\mathcal{S}$ is $softmax$ operation along dimension $j$. Each phase $k$ has its own $\{\mathbf{W}_\alpha^k, \mathbf{W}_\beta^k, \mathbf{W}_\varphi^k\}$ to increase network capacity.

\item\emph{Nonlinear mapping $ReLU(\mathcal{C} [\cdot , \cdot]) $}:
We use a learnable combination function $ReLU(\mathcal{C} [\mathbf{z}^k, \mathbf{v}^k]) $ to combine sparse local feature $\mathbf{z}^k$ and non-local feature $\mathbf{v}^k$ obtained from non-local block in Eq. \eqref{eq-11-1}. Finally, we defined our non-local operator $\mathcal{N}_k$ as:
\begin{equation}\label{nonlocal-out}
  \mathcal{N}_k (\overline{\mathbf{x}}^k) = ReLU(\mathcal{C}_k [\mathbf{z}^k, \mathcal{S}((\mathbf{W}_\alpha^k \mathbf{z}^k)^\mathrm{T} \mathbf{W}_\beta^k \mathbf{z}^k)\mathbf{W}_\varphi^k \mathbf{z}^k]).
\end{equation}

Here the input $\mathbf{z}^k$ corresponds to $\mathcal{G}_k(\overline{ \mathbf{x}}^k)$ in Eq. \eqref{sparse-3}, and $[\cdot, \cdot]$ denotes concatenation operator, $\mathcal{C}_k$ corresponds to a set of learnable weighted vectors which project the concatenated vector to a scalar (implemented as $1\times1$ convolution).
\end{itemize}
 As shown in Sec. \ref{Sec:extra}, the input $\mathbf{z}^k = \mathcal{G}_k(\overline{ \mathbf{x}}^k)$ for the non-local operator $\mathcal{N}_k$ has $7\times 7$ RF size.
 Therefore, our non-local operator can work without explicitly choosing patches. Moreover, the non-local operator $\mathcal{N}_k$ can take advantages of both local and non-local sparse features through learnable nonlinear combination, as depicted in Fig. \ref{fig-nonlocal}.

\subsection{Network Training} \label{sec:Training}
Our network comprises a cascade of $S$ EPN phases. Each phase corresponds to an iteration in the accelerated extra gradient algorithm in Eqs. \eqref{eq-17}-\eqref{eq-20} to learn a residual $\overline{{\mathbf{r}}}_{k}$, where $\overline{{\mathbf{r}}}_{k}$ is defined in Eq. \eqref{eq-26-0000}. The updating scheme is depicted in Fig. \ref{fig-1}. The learnable network parameters $\Theta= [\Theta_1, \cdots, \Theta_S]$, where $\Theta_k = [\mathcal{G}_k, \widetilde{\mathcal{G}}_k, \alpha_k, \beta_k, \varphi_k, \mathcal{C}_k, \gamma_k, \rho_k, \bm{\theta}_k]$ denotes the set of learnable parameters in $k^{th}$ phase. The training data are the pairs $\{\mathbf{y}_p, \mathbf{x}_p\}_{p=1}^P$, where $\mathbf{y}_p$ is the input raw data for reconstruction of compressive sensing and $\mathbf{x}_p$ is the corresponding ground truth image.
We design the loss function $\mathcal{L}(\Theta)$:
\begin{equation}\label{loss}
  \mathcal{L}(\Theta) = \frac{1}{PN}\sum_{p = 1}^P \|\hat{\mathbf{x}}_p(\mathbf{y}_p) - \mathbf{x}_p\|_2^2,
\end{equation}
where $\hat{\mathbf{x}}_p(\mathbf{y}_p)$ is the output of the network given input $\mathbf{y}_p$, $N$ is the size of image $\mathbf{x}_p$.

All the trainable parameters $\Theta$ are initialized using Xavier method \cite{Glorot10understandingthe} and trained on Adam solver \cite{kingma2014adam}. For testing result, we adopt the average Peak Signal-to-Noise Ratio (PSNR) as the measurement of evaluation for the performance of the network.

 \section{Experimental results}\label{sec:experiment}

 In this section, we examine the performance of our proposed EPN-Net in Compressive Sensing (CS) by comparing against the recent state-of-the-art methods.

This section is organized as following: we study the effectiveness of accelerated extra proximal-gradient inspired network without and with non-local block: EP-Net and EPN-Net respectively in Sec. \ref{compare-ista} and \ref{compare-nonlocal}.
  We summarize the comparisons of our proposed networks with several state-of-the-art image CS algorithms in Sec. \ref{sec:ista-comap}.
It has been shown in \cite{Zhang2018ISTANetIO} that ISTA-Net$^+$ outperforms several state-of-the-art image CS methods including both of optimization based and deep learning based approaches. Hence our focus is on the comparison with ISTA-Net$^+$. Our network is trained and tested in Tensorflow \cite{abadi2016tensorflow} on a machine with a GTX-1080Ti GPU.

 \textbf{Training and testing datasets:} For fair comparison, we use the same training and testing datasets as ISTA-Net$^+$ \cite{Zhang2018ISTANetIO}, 91 Images and Set11 \cite{kulkarni2016reconnet} respectively in each subsection.
 Here, we follow the same data preparation and result evaluation procedures as ISTA-Net$^+$. The ground truth $\{\mathbf{x}_p\}$ are $88,912$ patches with luminance components, which are randomly cropped into size $33\times 33$ from 91 Images dataset. The input $\mathbf{y}_p$ of data pair $\{ ( \mathbf{y}_p, \mathbf{x}_p)\}$ is generated by $\mathbf{y}_p = \mathbf{\Phi} \mathbf{x}_p$, where measurement matrix $\mathbf{\Phi}$ is  generated as a random Gaussian matrix with each row orthogonalized \cite{Zhang2018ISTANetIO}. The experiments are performed on CS ratios $10\%$ and $25\%$.

 On image CS problem, our network solves Eq. \eqref{eq-000} with $f(\mathbf{x}) = \|\mathbf{\Phi x} - \mathbf{y} \|^2_2$. It is solved iteratively as in the Eqs. \eqref{eq-23-2}-\eqref{eq-26-2} with the initial image $\mathbf{x}^0$. The initialization is the same as that in ISTA-Net$^+$.
That is to initialize $\mathbf{x}^0 = \mathbf{Q}_0 \mathbf{y}$, where $\mathbf{Q}_0$ is computed by $\mathbf{Q}_0= \mathop{\arg\min}_{\mathbf{Q}} {\| \mathbf{Q} \mathbf{Y} - \mathbf{X} \|^2_F} = \mathbf{X}\mathbf{Y}^\mathrm{T} (\mathbf{Y}\mathbf{Y}^\mathrm{T})^{-1}$, where $\mathbf{X} = [\mathbf{x}_1, \ldots, \mathbf{x}_P]$, $\mathbf{Y} = [\mathbf{y}_1, \ldots, \mathbf{y}_P]$ \cite{Zhang2018ISTANetIO}.
   \subsection{EP-Net vs. ISTA-Net$^+$}\label{compare-ista}


 In this subsection we compare the performance of the proposed EP-Net (without non-local operator) with ISTA-Net$^+$ \cite{Zhang2018ISTANetIO} for image CS reconstruction. Both networks exploit the sparsity of the solution in a transformed domain by minimizing $L_1$ norm. However, there are two major differences between the architectures of these two networks. Firstly, the ISTA-Net$^+$ is inspired on proximal gradient method, while the EP-Net relies on accelerated extra proximal gradient method. Secondly, in ISTA-Net$^+$ the residual $\mathbf{r}_k$ is considered as a linear function of $\mathbf{x}^k$, while in EP-net the $\mathbf{r}_k$ is designed as a nonlinear function of $\mathbf{x}^k$ to avoid inversion in ISTA-Net$^+$.
 For fair comparison, the number of channels in hidden layers $N_f$ is chosen the same for ISTA-Net$^+$ and EPN-Net: $N_f = 32$.

We compare ISTA-Net$^+$ and EPN-Net from two aspects: number of learnable parameters and accuracy on image compressive sensing reconstruction.

  \begin{itemize}
  \item\emph{Number of Learnable Parameters}: The number of learnable parameters in each phase of ISTA-Net$^+$ is $37442$ \cite{Zhang2018ISTANetIO}. The number of trainable parameters of each phase in EPN-Net is $\{ \mathcal{G}_k+ \widetilde{\mathcal{G}}_k + \gamma_k+\rho_k +\bm{\theta}_k = 32\times3\times3 \times(1+32\times2) + 32\times3\times3 \times(32\times2+1) + 1+2+32= 37475 \}$. Thus, the number of trainable parameters in each phase of  EPN-Net is almost the same to ISTA-Net$^+$.
  \begin{figure}[bp]
\begin{center}
   \includegraphics[width=0.5\linewidth]{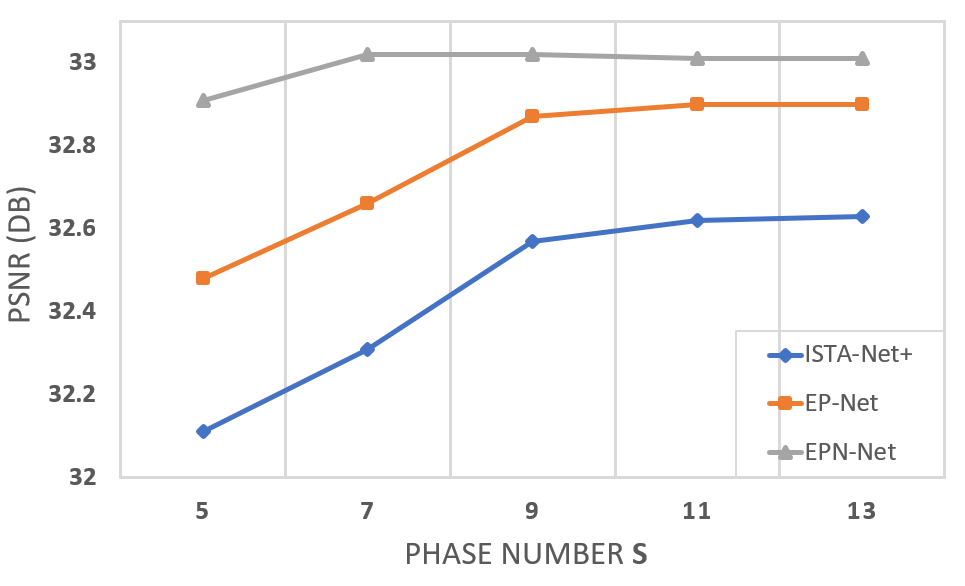}
\end{center}
   \caption{Average PSNR comparison between ISTA-Net$^+$, EP-Net and EPN-Net with various phase number on image compressive sensing problem on Set$11$ \cite{kulkarni2016reconnet} with CS ratio $25\%$.}
\label{fig-result-1}
\end{figure}

 \item\emph{Accuracy on Image Compressive Sensing Reconstruction}: We compare the reconstruction performance of EP-Net and ISTA-Net$^+$ with CS ratio $10\%$ and $25\%$, $9$-phase EPN-Net outperforms ISTA-Net$^+$ by $0.48$ dB and $0.30$ dB respectively, as shown in Tab. \ref{cs-result}. We also compare the reconstruction results of EP-Net and ISTA-Net$^+$ in a range of different phase number with CS ratio $25\%$, as shown in Fig. \ref{fig-result-1}. We observe that both PSNR curves increase with the increasing of phase number and approach to flat when phase number $S\geq9$.
     EP-Net achieves $0.3$ dB PSNR gain on average over ISTA-Net$^+$ at each phase number. To further demonstrate the superiority of accelerated extra proximal-gradient method over extending network depth, we compare $9$-phase EP-Net with $9$-phase ISTA-Net$^+$ and $15$-phase ISTA-Net$^+$, as shown in Tab. \ref{9vs1811}. Compared to $15$-phase ISTA-Net$^+$ which extends the depth of network by adding more layers, $9$-phase EP-Net achieves better accuracy ($0.27$ dB gain) using much shallower network and less learnable parameters.
\begin{table}[htp]
    \centering
    \caption{Compressive sensing reconstruction performance comparison of $9$-phase ISTA-Net$^+$, $15$-phase ISTA-Net$^+$, $9$-phase EP-Net and $7$-phase EPN-Net on Set11 \cite{kulkarni2016reconnet} with CS ratio $25\%$.}
    \setlength{\tabcolsep}{3mm}{
    \begin{tabular}{c|c|c}
        \hline\hline
        Algorithms & Learnable parameters &PSNR (dB) \\\hline
        $9$-phase ISTA-Net$^+$  &$336978$ & $32.57$ \\\hline
        $15$-phase ISTA-Net$^+$  &$561630 $& $32.60$ \\\hline
        $9$-phase EP-Net& $337275$ & $32.87$ \\\hline
        $7$-phase EPN-Net& $290997$ & $33.02$ \\\hline
    \end{tabular}}\label{9vs1811}
\end{table}
    \begin{table*}[bp]
    \centering
    \caption{Compressive sensing performance comparison on Set11 with CS ratio $25\%$. The performance is measured in terms of average PSNR (dB).  The left $5$ columns of the table is quoted from ISTA-Net$^+$ \cite{Zhang2018ISTANetIO}. Network-based methods are highlighted in bold. Here, we adopt EP-Net $9$ phases and EPN-Net $7$ phases.}
    \resizebox{\textwidth}{!}{
    \begin{tabular}{c|c|c|c|c|c|>{\columncolor[gray]{0.8}}c | >{\columncolor[gray]{0.8}}c}
        \hline\hline
        CS ratio &TVAL3 \cite{li2013efficient}&D-AMP \cite{metzler2016denoising}&\textbf{IRCNN} \cite{zhang2017learning}  &\textbf{ReconNet} \cite{kulkarni2016reconnet} &\textbf{ISTA-Net$^\mathbf{+}$} \cite{Zhang2018ISTANetIO}&\textbf{EP-Net} &\textbf{EPN-Net}\\\hline
        $10\%$ &$22.99$&$22.64$&$\mathbf{24.02}$&$\mathbf{24.28}$&$\mathbf{26.64}$&$\mathbf{27.12}$&$\mathbf{27.33}$\\\hline
        $25\%$ &$27.92$&$28.46$&$\mathbf{30.07}$&$\mathbf{25.60}$&$\mathbf{32.57}$&$\mathbf{32.87}$&$\mathbf{33.02}$\\\hline
    \end{tabular}}\label{cs-result}
\end{table*}
\end{itemize}

       \begin{figure}[htp]
       \subfigure{
       \label{fl-g} 
       \includegraphics[width=0.222\linewidth]{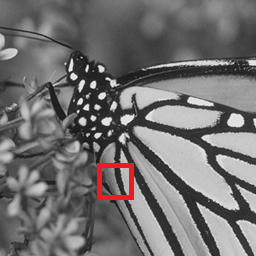}}
       \subfigure{
       \label{pa-g} 
       \includegraphics[width=0.222\linewidth]{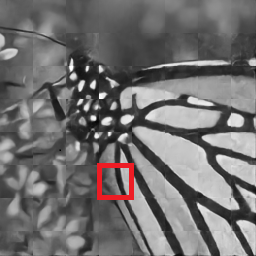}}
       \subfigure{
       \label{pa-ep} 
       \includegraphics[width=0.222\linewidth]{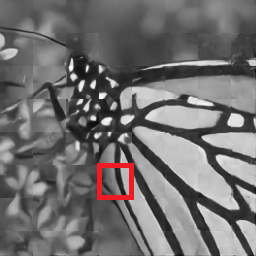}}
        \subfigure{
       \label{pa-epn} 
       \includegraphics[width=0.222\linewidth]{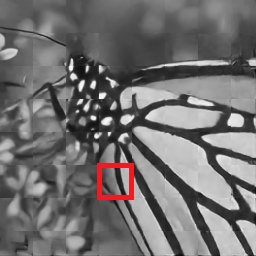}}
       \setcounter{subfigure}{0}
       \subfigure[Ground truth]{
       \label{bu-ep-2} 
       \includegraphics[width=0.223\linewidth]{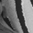}}       
       \subfigure[ISTA-Net$^+$ \protect \\ \ PSNR: $25.91$]{
       \label{bu-epn-2} 
       \includegraphics[width=0.223\linewidth]{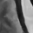}}
       \subfigure[EP-Net \protect \\ \ PSNR: $26.47$]{
       \label{bu-epn-2} 
       \includegraphics[width=0.223\linewidth]{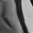}}
       \subfigure[EPN-Net \protect \\ \ PSNR: $26.58$]{
       \label{bu-epn-2} 
       \includegraphics[width=0.223\linewidth]{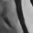}}
      \caption{ Reconstruction comparison of $9$-phase ISTA-Net$^+$, $9$-phase EP-Net and $7$-phase EPN-Net with CS ratio $10\%$, when applied to the Butterfly image in Set11 \cite{kulkarni2016reconnet}.}
      \label{inception22-2} 
      \end{figure}
 \begin{figure}[htp]
       \subfigure{
       \label{pa-g} 
       \includegraphics[width=0.222\linewidth]{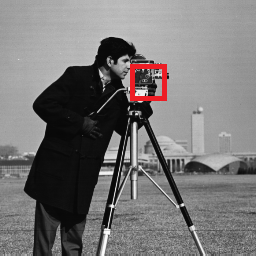}}
       \subfigure{
       \label{pa-g} 
       \includegraphics[width=0.222\linewidth]{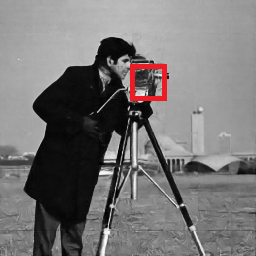}}
       \subfigure{
       \label{pa-ep} 
       \includegraphics[width=0.222\linewidth]{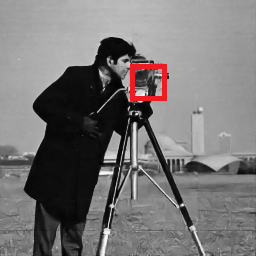}}
        \subfigure{
       \label{pa-epn} 
       \includegraphics[width=0.222\linewidth]{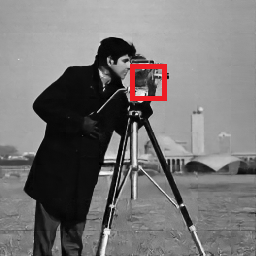}}
       \setcounter{subfigure}{0}
       \subfigure[Ground truth]{
       \label{bu-ep-2} 
       \includegraphics[width=0.223\linewidth]{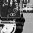}}       
       \subfigure[ISTA-Net$^+$ \protect \\ \ PSNR: $28.97$]{
       \label{bu-epn-2} 
       \includegraphics[width=0.223\linewidth]{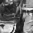}}
       \subfigure[EP-Net \protect \\ \ PSNR: $29.62$]{
       \label{bu-epn-4} 
       \includegraphics[width=0.223\linewidth]{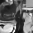}}
       \subfigure[EPN-Net \protect \\ \ PSNR: $29.73$]{
       \label{bu-epn-3} 
       \includegraphics[width=0.223\linewidth]{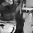}}
      \caption{ Reconstruction comparison of $9$-phase ISTA-Net$^+$, $9$-phase EP-Net and $7$-phase EPN-Net with CS ratio $25\%$, when applied to the Cameraman image in Set11 \cite{kulkarni2016reconnet}.}
      \label{inception22-3} 
      \end{figure}
 \begin{figure}[htp]
       \subfigure{
       \label{pa-g} 
       \includegraphics[width=0.222\linewidth]{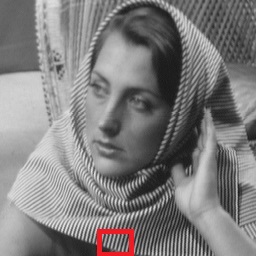}}
       \subfigure{
       \label{pa-g} 
       \includegraphics[width=0.222\linewidth]{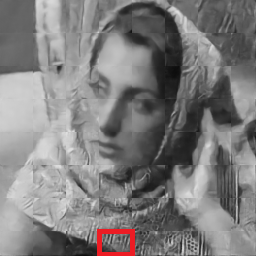}}
       \subfigure{
       \label{pa-ep} 
       \includegraphics[width=0.222\linewidth]{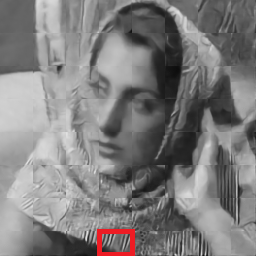}}
        \subfigure{
       \label{pa-epn} 
       \includegraphics[width=0.222\linewidth]{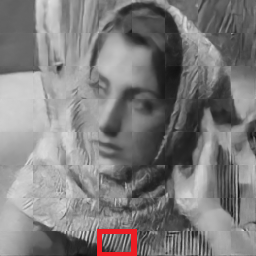}}
       \setcounter{subfigure}{0}
       \subfigure[Ground truth]{
       \label{bu-ep-2} 
       \includegraphics[width=0.223\linewidth]{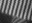}}       
       \subfigure[ISTA-Net$^+$ \protect \\ \ PSNR: $23.59$]{
       \label{bu-epn-2} 
       \includegraphics[width=0.223\linewidth]{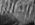}}
       \subfigure[EP-Net \protect \\ \ PSNR: $23.89$]{
       \label{bu-epn-4} 
       \includegraphics[width=0.223\linewidth]{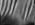}}
       \subfigure[EPN-Net \protect \\ \ PSNR: $24.27$]{
       \label{bu-epn-3} 
       \includegraphics[width=0.223\linewidth]{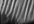}}
      \caption{ Reconstruction comparison of $9$-phase ISTA-Net$^+$, $9$-phase EP-Net and $7$-phase EPN-Net with CS ratio $10\%$, when applied to the Barbara image in Set11 \cite{kulkarni2016reconnet}.}
      \label{inception22} 
      \end{figure}
\subsection{EPN-Net vs. EP-Net}\label{compare-nonlocal}
 To demonstrate the effectiveness of non-local operator in EPN-Net, we compare EPN-Net and EP-Net from two aspects: number of learnable parameters and accuracy on image compressive sensing recnstruction.
    \begin{itemize}
  \item\emph{Number of Learnable Parameters}: The number of learnable parameters of each phase in EPN-Net is $\{ \mathcal{G}_k+ \widetilde{\mathcal{G}}_k + \alpha_k + \beta_k + \varphi_k+\mathcal{C}_k+\gamma_k+\rho_k +\bm{\theta}_k = 32\times3\times3 \times(1+32\times2) + 32\times3\times3 \times(32\times2+1)+ 32\times1\times1\times16 +32\times1\times1\times16+32\times1\times1\times32 + 64\times1\times1\times32+ 1+2+32= 41571 \}$. Thus, the number of trainable parameters in each phase of  EPN-Net is almost the same as EP-Net (with only $10.9\%$ increasing).
      However, note that EPN-Net tends stable after $7$ phases, while EP-Net needs $9$ phases. As shown in Tab. \ref{9vs1811}, $7$-phase EPN-Net, in fact, has less learnable parameters than $9$-phase EP-Net.
  \item\emph{Accuracy on Image Compressive Sensing Reconstruction}: We compare the reconstruction performance of EP-Net and EPN-Net with CS ratio $10\%$ and $25\%$, $7$-phase EPN-Net outperforms $9$-phase EP-Net by $0.21$ dB and $0.15$ dB respectively, as shown in Tab. \ref{cs-result}. We also compare the reconstruction results of EPN-Net and EP-Net in a range of different phase numbers with CS ratio $25\%$. The comparison results are shown in Fig. \ref{fig-result-1}. We observe that EPN-Net achieves average $0.2$ dB PSNR gain over EP-Net at each phase number. It is remarkable that the PSNR curve of EPN-Net gets flat at less phase number ($S = 7$). That illustrates that EPN-Net can employ shallower network without weakening performance. As shown in Tab. \ref{9vs1811}, EPN-Net with $7$ phases get $0.15$ dB PSNR gain with $13.7\%$ less parameters compared to EP-Net with $9$ phases.
\end{itemize}

Compared to state-of-the-art ISTA-Net$^+$ \cite{Zhang2018ISTANetIO}, both our proposed EP-Net and EPN-Net can get better reconstruction result with almost the same number of parameters. As visualized in Fig. \ref{inception22-2} to reconstruct Butterfly image with CS ratio $10\%$, $9$-phase EP-Net and $7$-phase EPN-Net can capture the inconspicuous contrast of the butterfly wings at the left-bottom part in the zoomed-in figures.
In Fig. \ref{inception22-3} to reconstruct Cameraman image with CS ratio $25\%$, the reconstructed images of $9$-phase EP-Net and $7$-phase EPN-Net have less flaw on background compared to ISTA-Net$^+$, such as the background region at the right-bottom of zoomed-in figures.
Fig. \ref{inception22} presents the reconstruction results for Barbara image in Set11 from ISTA-Net$^+$, $9$-phase EP-Net and $7$-phase EPN-Net with CS ratio $10\%$. We observe that the texture pattern of the scarf is better preserved by EPN-Net.
The better performance of EPN-Net illustrate the advantage of both extra proximal gradient algorithm inspired architecture and the use of non-local operator.


 \subsection{Comparison with State-of-the-Art Methods} \label{sec:ista-comap}
 We compare our proposed EP-Net and EPN-Net with the state-of-the-art image Compressive Sensing (CS) methods TVAL3 \cite{li2013efficient}, D-AMP \cite{metzler2016denoising}, IRCNN \cite{zhang2017learning},  ReconNet \cite{kulkarni2016reconnet} and ISTA-Net$^+$ \cite{Zhang2018ISTANetIO}, where the last three are network-based methods.
 Taking into account of the tradeoff between CS performance and network complexity, we adopt phase number $S = 9$ for EPN-Net and $S = 7$ for EP-Net when comparing with above methods.
 The reconstruction performance on Set11 dataset \cite{kulkarni2016reconnet} is shown in Tab. \ref{cs-result}. We observe that EP-Net and EPN-Net outperform all above state-of-the-art algorithms while EPN-Net achieves the best.

 \section{Conclusions and Future Work}\label{sec:conclusion}

 In this work we present a novel deep network EPN-Net to solve inverse problems in image reconstruction. The design of this network is inspired by our proposed algorithm in Sec. \ref{Sec:extra} that incorporates extra proximal-gradient algorithm and Nesterov's accelerated gradient algorithm. We use $L_1$-norm to enhance the sparsity of the solution in a learnable nonlinear transform domain.
 Moreover, we make full use of both local convolutional operation and non-local self-similarity exploiting operation to improve the accuracy of reconstruction.
 All parameters are discriminately learned through minimizing a loss function.
 Taking advantages of both optimization-based method and learnable network-based method, our network outperforms several existing state-of-the-art methods in image compressive sensing problems.

 Besides inverse imaging problems, one of our direction is to modify the proposed model to handle image segmentation and classification.
 Furthermore, our network adopts $L_1$-regularized prior, in the future we would like to try to replace $L_1$ norm by a non-convex approximation of $L_0$ norm, namely minimum concave penalty.

{\small
\bibliographystyle{ieee}
\bibliography{references}
}

\end{document}